\documentclass[3p,times,procedia]{elsarticle}
\usepackage{ecrc}
\usepackage[bookmarks=false]{hyperref}
    \hypersetup{colorlinks,
      linkcolor=blue,
      citecolor=blue,
      urlcolor=blue}
\usepackage{graphicx}
\usepackage{amsmath}
\usepackage{caption}

\volume{00}

\firstpage{1}

\journalname{Procedia Computer Science}

\runauth{Elena-Beatrice Nicola et al.}

\jid{procs}

\usepackage{amssymb}

\usepackage[figuresright]{rotating}

\begin{document}
\begin{frontmatter}



\dochead{28th International Conference on Knowledge-Based and Intelligent Information \& Engineering Systems (KES 2024)}%

\title{Investigating the Impact of Semi-Supervised Methods with Data Augmentation on Offensive Language Detection in Romanian Language}


\author[a]{Elena-Beatrice Nicola} 
\author[a]{Dumitru-Clementin Cercel\corref{cor1}}
\author[a,b,c]{Florin	Pop}

\address[a]{Faculty of Automatic Control and Computers, National University of Science and Technology POLITEHNICA Bucharest, Romania}
\address[b]{National Institute for Research and Development in Informatics (ICI), Bucharest, Romania}
\address[c]{Academy of Romanian Scientists, Bucharest, Romania}

\begin{abstract}    
Offensive language detection is a crucial task in today’s digital landscape, where online platforms grapple with maintaining a respectful and inclusive environment. However, building robust offensive language detection models requires large amounts of labeled data, which can be expensive and time-consuming to obtain. Semi-supervised learning offers a feasible solution by utilizing labeled and unlabeled data to create more accurate and robust models. In this paper, we explore a few different semi-supervised methods, as well as data augmentation techniques. Concretely, we implemented eight semi-supervised methods and ran experiments for them using only the available data in the RO-Offense dataset and applying five augmentation techniques before feeding the data to the models. Experimental results demonstrate that some of them benefit more from augmentations than others.
\end{abstract}

\cortext[cor1]{Corresponding author.}
\end{frontmatter}

\email{dumitru.cercel@upb.ro}


\section{Introduction}
The problem of detecting offensive language in social media poses quite a challenge, as offensive phrases can be implicit or not \cite{guo2023implicit}. According to the definition \cite{collins} of the offensive term, it is described as something that, due to its rudeness or offensive nature, irritates or embarrasses individuals. As such, it is clear that the type of language we are trying to detect can be expressed in a variety of ways.
However, a good offensive language detector should be robust enough to detect offensive phrases regardless of their target or type of content. For example, we should be able to detect offensive language if it targets one's religion, sex, or just plain insults a person. This raises the question of whether it is possible to create such a detector.

Another aspect of social media platforms is their international character. Content can be found in various languages on all platforms, meaning that offensive language should be detected in all of them. One language of particular interest to us is Romanian, prompting the question of the best approach to detecting offensive language in Romanian.

In this paper, we set about answering these questions by testing a few approaches. Because adequate labeled data is essential for good performance and generalization, but obtaining these labels can be difficult, time-consuming, and expensive, we decided to explore the performance of eight different semi-supervised methods. Another reason is that fully supervised approaches have been tested before \cite{hoefels2022coroseof,cojocaru2022news,trandabat2022detecting, paraschiv2022fighting}.
We ran experiments using FixMatch \cite{sohn2020fixmatch} with and without contrastive regularization (CR) \cite{lee2022cr}, FreeMatch \cite{Wang2022freematch}, MixMatch \cite{berthelot2019mixmatch}, mean teacher (MeanTeacher) \cite{tarvainen2017mean}, noisy student (NoisyStudent) \cite{park2020noisy}, label propagation (LabelPropagation) \cite{zhul2002label}, and semi-supervised generative adversarial network (SGAN) \citep{odena2016semi,su2023ssl}.


Furthermore, we use five data augmentation techniques to see how they would affect the performance of the above-mentioned methods. These techniques are: paraphrasing using a T5 model \cite{t5rohugg} we fine-tuned to paraphrase Romanian sentences, easy data augmentation (EDA) \cite{wei2019eda}, sentence generation using a GPT2-large model \cite{niculescu2021rogpt2} we fine-tuned to generate offensive language in Romanian, Manifold Mixup \cite{verma2019manifold}, and back-translation with the following flow: Romanian-English-Italian-English-Romanian. Most of them led to significant improvements. 

The main contributions of this work are as follows:
(i) implementing and running experiments using semi-supervised methods on detecting offensive language in Romanian;
(ii) implementing augmentations and running experiments to enhance the performance of the semi-supervised methods on detecting offensive language in Romanian; and
(iii) making a comparison between the obtained results and concluding what is the best approach.
    

\section{Related Work}

Researchers have previously approached the task of detecting offensive language in Romanian. \citet{paraschiv2022fighting} have introduced the dataset used in this work. They have also tested a few supervised methods, such as SVM, gradient boosting, CNNs, transformer-based architectures, and vocabulary graph convolutional networks (VGCNs) \cite{lu2020vgcn}. Their results with the Romanian BERT (RoBERT) model \cite{masala2020robert} are micro-F1 of 79.40\% and macro-F1 of 78.15\%. They obtained the best results using the VGCN-enhanced RoBERT technique, with micro-F1 of 83.31\% and macro-F1 of 82.53\%. 

\citet{cojocaru2022news} introduced another dataset on Romanian offensive language with the comments extracted from a news website. This dataset contains 4,052 samples labeled as Non-offensive, Targeted insult, Racist, Homophobic, and Sexist. They experimented with a few supervised classification methods, such as SVM, RoBERT, and multilingual BERT \cite{pires2019multilingual}. The best results were obtained with the RoBERT approach, which resulted in an F1-score of 0.74\%. 

\citet{trandabat2022detecting} have also approached the problem of detecting offensive language in Romanian. They also introduced a dataset, as well as a few supervised classification methods. Their dataset contains 24,000 samples divided into offensive and non-offensive classes. The comments were extracted from YouTube, Facebook, and Twitch. The classification methods are Naive Bayes, SVM, passive-aggressive, and logistic regression. Their best results are obtained using SVM with an F1-score of 92.09\%.

\section{Method}

\subsection{Semi-Supervised Methods}



This work explores several semi-supervised methods for training deep learning classifiers as follows. \textbf{FixMatch} is implemented as described by \citet{sohn2020fixmatch}. We have two distinct loss terms, one for the labeled part and another for the labeled part. They both use the cross entropy function, but with differently computed inputs. The labeled loss is applied to weakly augmented input samples. The unlabeled loss uses two variations of the input sample obtained through data augmentation into two strength levels, with the weaker one being the pseudo-label. Both transformed results are forwarded through the model to get a probability distribution over the possible class labels. 

The idea behind \textbf{contrastive learning} \cite{balestriero2023cookbook} is to create augmented versions of input samples and treat each of them as a positive pair with its original sample. These positive pairs are then contrasted with other negative samples in the dataset. \citet{lee2022cr} integrated the method within the semi-supervised paradigm by introducing a contrastive regularization loss term that uses pseudo-label information for the unlabeled data; this version is used in this work.

The core idea of \textbf{FreeMatch} \cite{Wang2022freematch} is to modify the confidence threshold for pseudo-labeling dynamically. This is done through consistency regularization and self-training using labeled and high-confidence pseudo-labeled data.

\textbf{MixMatch} \cite{berthelot2019mixmatch} incorporates components from both pseudo-labeling and consistency regularization. The unlabeled data is assigned corresponding pseudo-labels and, together with the labeled data, is blended into novel entries using the mixup technique \cite{zhang2017mixup}. This method applies linear interpolation, and the objective enforced on the trained model is that the prediction of the interpolated inputs should be the interpolation of the individual predictions.

\textbf{MeanTeacher} \cite{tarvainen2017mean} improves temporal ensembling \cite{laine2016temporal} by maintaining an exponential moving average of the model weights, allowing more frequent target updates. The student should produce predictions that are close to the real target as well as the teacher's proposal. The balance between the two opposing forces can be controlled algorithmically by assigning different weights for the two corresponding loss terms or by keeping unit weight for the classification cost and adjusting only the consistency cost’s weight.

\textbf{NoisyStudent} \cite{park2020noisy} implies an iterative algorithm that assigns pseudo-labels to unlabeled data and trains, in a supervised manner, a classifier on the concatenation of labeled and pseudo-labeled data. Initially, a model called teacher is fitted to the available labeled data. With the accumulated knowledge, the first teacher provides pseudo-labels on the unlabeled data. A different model, the student, is trained on all the data. Then, it becomes the next teacher for a new student, and the process repeats for three iterations in this work. The key element added by NoisyStudent is the insertion of noise during student training and not during teacher training.

\textbf{LabelPropagation} \cite{zhul2002label} involves the construction of a similarity graph between the available input samples. For several epochs, as in our case, five, a sequence of steps is strictly followed. After feature extraction and graph construction, the complete graph is reduced to the nearest neighbor graph of a specific order to make the computation tractable and prune unnecessarily minor similarities. The final loss function is weighted accordingly with sample weights to account for various issues that may hinder the learning process, such as class imbalance or overconfidence.

The \textbf{SGAN} model consists of the following three parts: a text encoder, a generator, and a discriminator, as described in the literature \citep{odena2016semi,su2023ssl}. The encoder extracts sentence-level features from both labeled and unlabeled data, along with an extra linear layer stack that compresses the sentence-level features. The generator transforms the random noise into sentence-level features. At the same time, the discriminator classifies the sentence-level embedding into one of five categories: Other, Abuse, Insult, Profanity, or Fake (which means that the generator has generated it).


\subsection{Data Augmentation Strategies}

At least half of the tested methods rely on enforcing consistency regularization, which requires performing data augmentation to generate artificial examples similar to the real ones at varying degrees of similarity. Thus, the augmentation performed in the simple case adds Gaussian noise of mean 0 to the embeddings, which can be controlled to be weak or strong by adjusting the magnitude of the standard deviation. 
Furthermore, we have tried a few more complex augmentation methods to increase the number of sentences in the dataset and improve the performance of the models. These methods include paraphrasing, EDA, sentence generation using the RoGPT2-large, MixUp, and back-translation.

For \textbf{paraphrasing}, the T5 model \cite{t5rohugg} pre-trained on the Romanian language was fine-tuned using two datasets: RO-STS \cite{dumitrescu2021liro} and paraphrase-ro\footnote{\url{https://huggingface.co/datasets/BlackKakapo/paraphrase-ro}}. Then, the model was applied to every sample in the training part of the RO-Offense dataset, doubling the training entries available. After that, we applied the semi-supervised methods using the enhanced dataset, trained the corresponding models, and evaluated them on the task of classifying offensive language. 

\textbf{EDA} was implemented as described by \citet{wei2019eda}. As the training dataset consists of almost 10k sentences, we used an $\alpha$ of 0.1 and generated four augmentations per sentence. 

For generating new samples, we used the \textbf{RoGPT2-large} transformer architecture \cite{niculescu2021rogpt2} pre-trained on the Romanian language, and we fine-tuned it using the RO-Offense dataset to generate offensive language. We generated 100k new sentences that we then used to train the semi-supervised models. 

In this work, \textbf{Manifold Mixup} \cite{verma2019manifold} is used. It is a newer version of the original Input Mixup \cite{zhang2017mixup}, which linearly interpolates the hidden states of two distinct data samples along with their corresponding classes. Following \citet{chowdhury2020cross}, first, a mixup ratio $\lambda$ is sampled from a $\beta$ distribution: $\lambda \sim \beta(\alpha, \alpha)$. Next, a hidden layer $l$ is randomly chosen for the mixup. Let $h^l_i$ and $h^l_j$ be the randomly chosen, $l^{th}$ hidden layer output from the $i^{th}$, and $j^{th}$ comment samples. Then, the two outputs can be mixed up as: $\Tilde{h}^l_i = \lambda \times h^l_i + (1 - \lambda) \times h^l_j$, with $\Tilde{h}^l_i$ being the augmented hidden state.
The same $\lambda$ is used to mix not just the hidden states of the data samples $i$ and $j$ but also the corresponding ground truth labels for each class $k$ in the RO-Offense dataset.

A few models from Helsinki-NLP\footnote{\url{https://huggingface.co/Helsinki-NLP}} were used for \textbf{back-translation}. Firstly, the\textit{ opus-mt-roa-en} model was used to translate from Romanian to English, then \textit{opus-mt-en-it} to translate from English to Italian. Next, we used the \textit{opus-mt-it-en} to translate back to English, and finally, the \textit{opus-mt-en-ro} was used to get back to Romanian. We did that because we felt that translating to English and back wouldn't change the sentence enough.


\section{Experimental Setup}

\subsection{Dataset}

Our experiments are based on the RO-Offense dataset\footnote{\url{https://github.com/readerbench/ro-offense}} \cite{paraschiv2022fighting}, consisting of two parts. The first part is a labeled collection of 12,445 comments extracted from a Romanian sports website, with 7,873 offensive samples and 4,572 non-offensive ones. Besides these samples already split in a fixed training and test set, an additional validation set of 1,991 samples was used for hyperparameter tuning. The second part is similar to the first one, containing comments from the same Romanian sports website, but all the data is unlabeled. This part includes 65,000 comments.

The comments are distributed across four classes, namely ABUSE, INSULT, PROFANITY, and OTHER (i.e., the label for non-offensive comments). The counts for each class in each dataset split can be observed in Table \ref{tab1:offense split}. With minor variations across the three splits, the identified categorical distribution is as follows: OTHER – 36\%, ABUSE – 28\%, INSULT – 23\%, and PROFANITY 13\%. Thus, the imbalance ratio (the proportion of the most frequent class divided by the proportion of the least frequent one) is approximately 2.77. This value is relevant because the studied methods in this work perform best when the available data is not highly imbalanced, preferably wholly balanced. Several works \citep{lee2021abc,hyun2020class} outlined the impact of various degrees of imbalance on the final performance of semi-supervised learning methods. What can be inferred from their analysis is that a noticeable degradation happens for an imbalance ratio much higher than the one associated with RO-Offense, from at least 10. Therefore, no special class imbalance handling must be implemented to achieve top results.

As observed by ~\citet{paraschiv2022fighting}, given the specificity of the dataset with the comments being exclusively from the sports domain, there has been observed a bias towards offensiveness that is introduced by mentions of certain entities, such as organizations, people, and teams. To account for this observation and to avoid the situation in which a classifier would overfit on these features, an anonymized version of the dataset was created, where the previously mentioned entities are marked with generic tags ([ORG] for organization and [PERS] for person). The anonymized version was used in all the experiments performed in the current work, and the entity tags were filtered out.


\subsection{Text Preprocessing} 
\begin{figure}
    \includegraphics[width=1\linewidth]{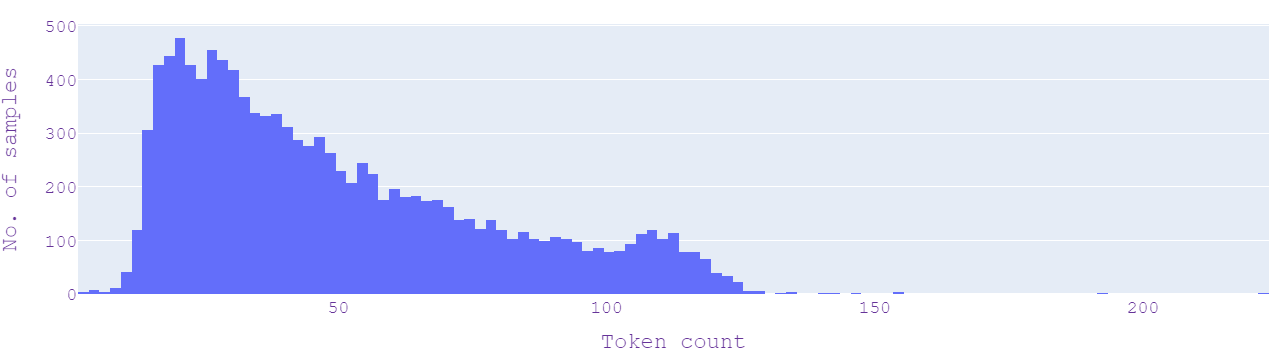}
    \caption{Length distribution in the anonymized RO-Offense training set after preprocessing and tokenization.}
    \label{fig3}
\label{tab1:length_distribution}
\end{figure}

All the dataset splits follow a similar token distribution to the one presented in Figure \ref{tab1:length_distribution}. A tokenization step is required in the preprocessing stage to use the investigated models. All the samples are brought to the same length, either by padding or truncating, by employing the functionalities of the corresponding tokenizer from Hugging Face’s transformer library \cite{wolf2019huggingface}. The final length was chosen to be 96 tokens. This determines a balanced trade-off between keeping enough information for each sample and the computational requirements of training the model.

In the first stage of the text processing flow, we perform Unicode normalization (i.e., diacritics removal), lowercasing, entity tag removal, deduplicating whitespaces, and removal of repeated letters. Then, the normalized text is passed to the tokenize stage, which applies the Hugging Face tokenizer that matches the model used later in the pipeline. This generates an iterable for token ids as well as the attention mask for each instance. When padding is applied, the added padding tokens will correspond to zero values in this mask.

Since the unlabeled part isn’t anonymized in the dataset, a few more steps are needed, and we had to do this before applying the same steps as for the labeled part. Fortunately, the dataset provides a list of all the persons and organizations encountered in the comments, which reduces the task of anonymizing the samples and makes it easier to identify if there are persons or organizations mentioned in the comment and replace them with the corresponding tag of [PERS] or [ORG].

\begin{table}
\captionsetup{justification=centering}
\caption{The number of comments for each RO-Offense split.}
\label{tab1}
\centering
\begin{tabular}{l|lll|l}
\hline
Class  & Training & Validation & Test & Total \\
\hline
OTHER & 3,649 & 720 & 898 & 5,267 \\
ABUSE & 2,760 & 570 & 684 & 4,014 \\
INSULT & 2,242 & 435 & 579 & 3,256 \\
PROFANITY & 1,294 & 261 & 331 & 1,886 \\
\hline
\end{tabular}
\label{tab1:offense split}
\end{table} 

\subsection{Baseline Models}

Several alternatives for the model architecture were considered. The best of them was chosen as the baseline for the rest of the experiments, which involved semi-supervised training.
The results are outlined in Table \ref{tab1:baseline}. The best configuration for the baseline involves the RoBERT transformer and a multilayer perceptron (MLP) classifying head. This configuration outperformed the multilingual approach, XLM-R~\cite{conneau2019unsupervised}, because the targeted masked language modeling allows the model to focus on only one language. It also outperformed the approach based on the randomly initialized embedding layer because of the available information learned during pre-training.

\begin{table}
\caption{A selection of the best-performing configurations for each encoder alternative. RoBERT* refers to the RoBERT architecture with randomly initialized embeddings.}
\label{tab2}
\centering
\begin{tabular}{l|l|l|l|l|l}
\hline
Model & Accuracy & Precision & Recall & Micro-F1 & Macro-F1 \\
\hline
\small{RoBERT + BiGRU} \cite{paraschiv2023offensive} & - & \textbf{81.64} &  79.36 & 80.48 & 79.32 \\
\small{RoBERT* + MLP} & 69.79 & 72.78 & 66.56 & 69.44 & 68.53 \\
\small{RoBERT + MLP}  & \textbf{81.00} & 81.57 & \textbf{80.56} & \textbf{81.05} & \textbf{80.23}\\
\small{XLM-R + MLP} & 66.54 & 76.00 & 54.43 & 63.45 & 62.99 \\
\hline
\end{tabular}
\label{tab1:baseline}
\end{table}



\subsection{Hyperparameters}

The text classifiers have been trained as follows: 16 epochs for FixMatch, 20 epochs for MeanTeacher, and 15 epochs for FixMatch + CR, FreeMatch, and MixMatch. The batch size is 16 for all the semi-supervised methods except FreeMatch, for which it is 32. 
The optimizer is AdamW for all the models, though it has different parameters. For FixMatch, FixMatch + CR, FreeMatch, MixMatch, MeanTeacher, NoisyStudent, and LabelPropagation, the optimizer parameters are weight decay of 1e-3, learning rate of 5e-3 for the multilayer perceptron on top, and no weight decay. The learning rate for the RoBERT encoder is 1e-5.
The best configuration of MixMatch, MeanTeacher, and NoisyStudent establishes standard Gaussian noise and dropout with a rate of 0.2 as noise-adding methods when not using any of the data augmentation strategies.

\begin{figure}
    \includegraphics[width=1\linewidth]{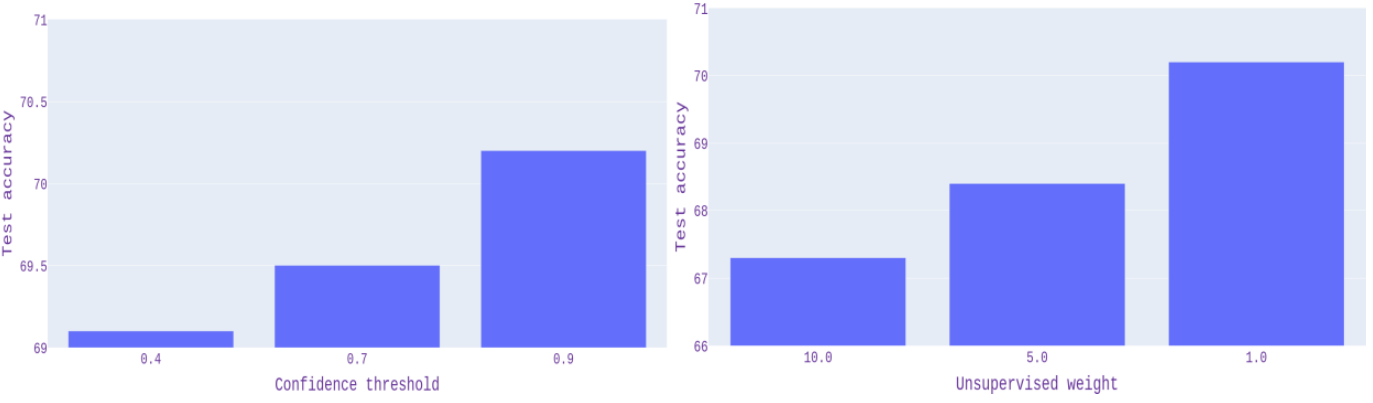}
    \caption{Results for hyperparameters of FixMatch. A higher confidence threshold produces marginally better results. Unsupervised weight should assign equal contributions to supervised and unsupervised components.}
    \label{tab1:fig3}
\end{figure}

For FixMatch and all other training approaches, the transformer encoder must be fine-tuned at a lower learning rate to avoid destroying the pre-training knowledge. Moreover, weight decay was found to be detrimental to the learning process of the same encoder, so it was disabled for this component of the model.
The best configuration of hyperparameters is characterized by a confidence threshold of 0.9 and an unsupervised weight of 1.0 (equal importance for supervised loss and unsupervised loss components).
A comparative study has been performed on two hyperparameters of the FixMatch method, namely the confidence threshold and the unsupervised weight. According to the outline presented in Figure \ref{tab1:fig3}, a larger confidence threshold is desired because it avoids including low-confidence examples in the early stages of learning that may affect the overall process. As training progresses, the model will be more confident in its prediction for more difficult samples and gradually leverage most unlabeled samples. Regarding the unsupervised weight, if a static value is chosen, it should assign an almost equal contribution to the two loss terms. This avoids relying too much on the pseudo-labels of the unlabeled data at the beginning of training. If an updated schedule is chosen, the weight value should start with a small value and increase gradually.

For contrastive regularization, there is an addition: the architecture includes a projection network optimized using AdamW with no weight decay and a learning rate of 1e-3. The projection network takes the RoBERT embeddings and projects them into a lower-dimensional space where pairwise similarities will be computed and used during training. The best configuration of contrastive regularization assigns equal weights to the three loss components: supervised, unsupervised, and contrastive. Similar to FixMatch, threshold values are required to use only high-confidence samples. contrastive regularization’s best configuration has equal threshold values of 0.7 for both the unsupervised and contrastive components.

The best configuration of hyperparameters corresponding to the FreeMatch method is characterized by a fairness weight of 0.01 and an unsupervised weight of 1.0.

The unsupervised weight in MixMatch is 4, which translates to an effective weight of 1 because the approach divides the parameter’s value by the number of classes.
Also, following \citet{chen2020MixText}, we set the parameter $\alpha$ to 0.3. This is justified because, in a high-label setting, where enough labeled data guides the model, the augmentations can be more intense without the risk of affecting the learning process.

For MeanTeacher, the moving average of the model weights is updated using a coefficient of 0.99. The unsupervised weight isn't constant throughout training, as it follows the schedule proposed in the original description \cite{tarvainen2017mean}.

For NoisyStudent, the text classifier has been trained for 20 initial supervised epochs, where the teacher learns to discriminate the available labeled data, and then for three iterations of 20 epochs each of student training.
A comparative study has been performed on the impact of adding noise during student training. It was observed that the performance decreases only marginally.

\begin{figure}
    \includegraphics[width=1\linewidth]{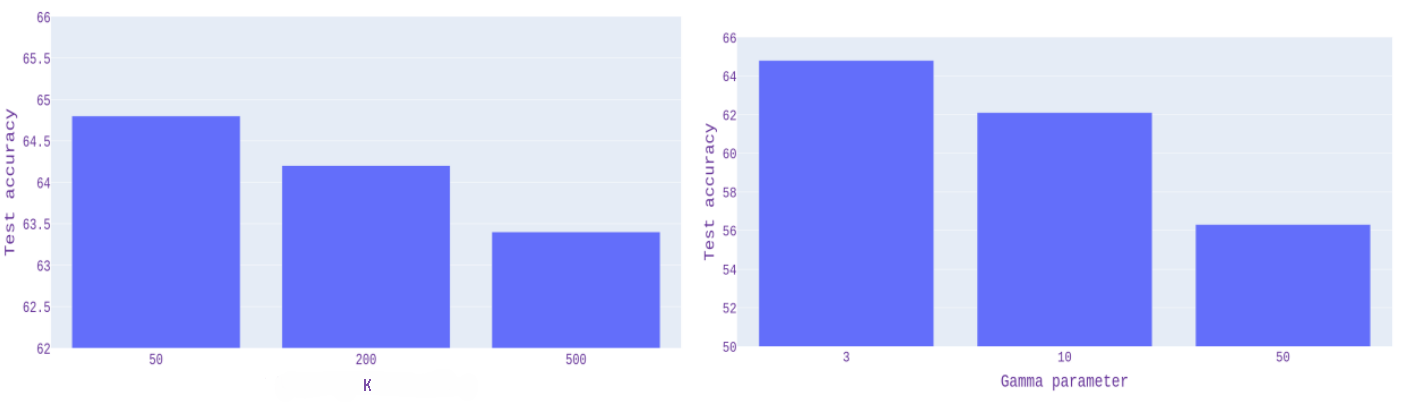}
    \caption{Results for hyperparameters of LabelPropagation: $k$ and $\gamma$.}
    \label{tab1:fig4}
\end{figure}

The text classifier has been trained using the LabelPropagation approach for five initial fully supervised epochs and an additional five epochs in which the actual propagation is applied.
The best configuration of hyperparameters is characterized by a value of 50 for $k$, 0.99 for $\alpha$, and 3 for $\gamma$. For larger values of $k$, the performance has marginally decreased and at the cost of reduced sparsity, introducing more computational overhead (see Figure \ref{tab1:fig4}). Regarding the influence of the $\gamma$ parameter (see Figure \ref{tab1:fig4}), larger values scale the elements of the affinity matrix in such a way that all entries will be closer to 0 because they are the result of dot products between normalized feature vectors so that they will be situated in the interval [-1, 1]. This has an opposing effect relative to the neighborhood size $k$, promoting sparsity inside the affinity matrix.

The SGAN has been trained for 40 epochs using the AdamW optimizer as before but with a weight decay and learning rate of 1e-3 for both the generator and the discriminator.
The best results were obtained using the RoBERT as the text encoder, with each linear layer followed by a LeakyReLU activation function layer and a dropout one.

\section{Results}

As shown in Table \ref{tab1:fig8}, most evaluated methods have improved over the baseline supervised approach, even without any augmentations. Only MeanTeacher and SGAN registered a decrease in performance when comparing macro-F1 scores. The most significant decrease was in the case of SGAN, with 8.07\%, while the highest improvement is displayed by FixMatch, with a 19.15\% increase in the macro-F1 score.

FixMatch also displays improvements when using almost all of the augmentation techniques. Significant performance improvement was obtained with the generative data augmentation using RoGPT2, with a 7.40\% increase in macro-F1 score. Though the precision is the best when using back-translation, the F1 score has not improved as much as in the case of generative augmentation. The only strategy that did not lead to improvements was when using EDA.

Note that the introduction of contrastive regularization alongside  FixMatch (i.e., FixMatch + CR) didn't bring any improvements. In all cases, with or without any augmentation, it was always behind the corresponding case of simple FixMatch by approximately 2.8\%.

FreeMatch improved over the baseline, with a 15.16\% gain in the macro-F1 score. Again, it benefited from all of the augmentation strategies except EDA. The best results were again obtained when employing generative augmentation, with an increase of 5.80\% for macro-F1.

MixMatch resulted in more of the same, though this time, the improvement when using back-translation was slight, with only a 0.35\% increase for macro-F1.

However, MeanTeacher registered the biggest increase when using back-translation, with a 4.04\% gain in macro-F1. The rest of the augmentation strategies showed minor improvements, with a slight decrease when using EDA.

NoisyStudent is the first method that has registered improvements for all the augmentation strategies. Again, the most beneficial one was generative augmentation, with a gain of 11.30\% in macro-F1.

LabelPropagation is different from the other methods because it has shown a slight decrease of 0.04\% for macro-F1 when using back-translation. The biggest gain was also made when using generative augmentation, with 3.50\% for macro-F1.

SGAN is another method that has registered improvements for all the augmentation methods. This time, the most significant gains were made when using back-translation, with 4.30\% in macro-F1. Despite this, it can be observed that SGAN has the worst performance among the collection of the tested semi-supervised methods. This is likely because of the more complex setup required to have a GAN trained in a semi-supervised method and by the added complexity of not only classifying the comments but also figuring out which are real and which have been generated by the generator network.

\begin{table}
\caption{Results for best-performing configurations for each semi-supervised method against our best-supervised approach and the best-supervised approach by \cite{paraschiv2022fighting}.}
\label{tab3}
\centering
\begin{tabular}{l|l|l|l|l|l|l}
\hline
Method & Augmentation & Accuracy & Precision & Recall & Micro-F1 & Macro-F1 \\
\hline
RoBERT + CNN~\cite{paraschiv2022fighting} & - & 79.10 & 81.09 & 77.78 & 79.40 & 78.15 \\
RoBERT + MLP & - & 81.00 & 81.57 & 80.56 & 81.05 & 80.23 \\
\hline
FixMatch & - & 81.35 & 82.16 & 79.97 & 80.98 & 78.89 \\
FixMatch & Paraphrase & 83.70 & 83.12 & 81.73 & 82.53 & 80.42 \\
FixMatch & EDA & 80.96 & 81.95 & 76.56 & 80.65 & 77.93 \\
FixMatch & RoGPT2-large & \textbf{91.25} & 84.96 & \textbf{86.57} & 85.76 & \textbf{86.29} \\
FixMatch & Manifold Mixup & 83.62 & 84.45 & 82.22 & 83.29 & 81.12 \\
FixMatch & Back-translation & 88.27 & \textbf{92.25} & 82.45 & 83.51 & 79.56 \\
\hline
FixMatch + CR & - & 78.69 & 78.58 & 76.17 & 77.36 & 76.35 \\
FixMatch + CR & Paraphrase & 79.97 & 80.64 & 77.81 & 78.95 & 77.97 \\
FixMatch + CR & EDA & 79.37 & 77.88 & 76.82 & 77.40 & 75.74 \\
FixMatch + CR & RoGPT2-large & 85.59 & 88.58 & 80.17 & 84.36 & 82.65 \\
FixMatch + CR & Manifold Mixup & 80.99 & 80.82 & 78.45 & 79.62 & 78.71 \\
FixMatch + CR & Back-translation & 80.74 & 79.89 & 77.38 & 77.52 & 76.56 \\
\hline
FreeMatch & - & 78.64 & 76.33 & 76.22 & 76.31 & 74.90 \\
FreeMatch & Paraphrase & 80.72 & 77.93 & 77.29 & 78.38 & 76.62 \\
FreeMatch & EDA & 78.33 & 76.86 & 76.00 & 76.18 & 74.53 \\
FreeMatch & RoGPT2-large & 88.54 & 84.83 & 83.62 & 84.22 & 80.70 \\
FreeMatch & Manifold Mixup & 80.82 & 78.43 & 78.34 & 78.39 & 77.04 \\
FreeMatch & Back-translation & 78.72 & 76.37 & 76.03 & 76.34 & 75.08 \\
\hline
MixMatch & - & 79.40 & 79.29 & 78.51 & 78.88 & 78.53 \\
MixMatch & Paraphrase & 80.54 & 80.51 & 80.14 & 79.89 & 80.59 \\
MixMatch & EDA & 79.53 & 79.75 & 78.88 & 79.03 & 78.31 \\
MixMatch & RoGPT2-large & 90.80 & 90.49 & 83.51 & \textbf{86.86} & 86.03 \\
MixMatch & Manifold Mixup & 81.46 & 81.42 & 80.58 & 80.99 & 80.62 \\
MixMatch & Back-translation & 79.33 & 79.33 & 78.30 & 78.93 & 78.88 \\
\hline
MeanTeacher & - & 69.67 & 78.84 & 54.66 & 64.25 & 59.35 \\
MeanTeacher & Paraphrase & 71.44 & 80.93 & 56.38 & 66.44 & 60.90 \\
MeanTeacher & EDA & 69.57 & 79.16 & 54.35 & 64.09 & 59.00 \\
MeanTeacher & RoGPT2-large & 78.37 & 78.94 & 58.56 & 67.24 & 60.25 \\
MeanTeacher & Manifold Mixup & 71.94 & 80.94 & 56.51 & 64.87 & 60.69 \\
MeanTeacher & Back-translation & 88.89 & 89.92 & 57.55 & 66.49 & 63.39 \\
\hline
NoisyStudent & - & 69.46 & 73.47 & 62.83 & 67.58 & 63.82 \\
NoisyStudent & Paraphrase & 71.56 & 74,65 & 63.85 & 69.71 & 65.33 \\
NoisyStudent & EDA & 69.14 & 73.31 & 62.44 & 67.63 & 64.27 \\
NoisyStudent & RoGPT2-large & 81.06 & 84.87 & 71.43 & 77.64 & 75.12 \\
NoisyStudent & Manifold Mixup & 71.06 & 75.28 & 64.24 & 69.54 & 65.66 \\
NoisyStudent & Back-translation & 71.94 & 75.41 & 65.59 & 69.19 & 65.73 \\
\hline
LabelPropagation & - & 79.89 & 78.61 & 78.30 & 78.51 & 77.62 \\
LabelPropagation & Paraphrase & 82.23 & 80.48 & 79.61 & 80.11 & 78.85 \\
LabelPropagation & EDA & 79.53 & 78.69 & 78.29 & 78.35 & 77.75 \\
LabelPropagation & RoGPT2-large & 82.79 & 85.81 & 83.50 & 84.64 & 81.12 \\
LabelPropagation & Manifold Mixup & 81.42 & 79.97 & 79.78 & 80.06 & 78.98 \\
LabelPropagation & Back-translation & 80.93 & 81.08 & 80.20 & 78.85 & 77.58 \\
\hline
SGAN & - & 68.73 & 79.44 & 56.15 & 65.66 & 51.67 \\
SGAN & Paraphrase & 70.76 & 80.73 & 57.84 & 66.75 & 53.11 \\
SGAN & EDA & 68.94 & 79.28 & 56.14 & 66.10 & 52.04 \\
SGAN & RoGPT2-large & 71.73 & 79.74 & 60.75 & 68.96 & 52.77 \\
SGAN & Manifold Mixup & 70.92 & 81.45 & 58.00 & 67.53 & 53.50 \\
SGAN & Back-translation & 72.17 & 81.65 & 58.82 & 68.11 & 55.97 \\
\hline
\end{tabular}
\label{tab1:fig8}
\end{table}

When applying the augmentation techniques, the best results were yielded by generating new samples using RoGPT2-large. This technique resulted in an average increase of 6\% in F1 score. The biggest gain was registered by the NoisyStudent method, with 11.3\% in the macro-F1 score. The lowest increase was registered by both MeanTeacher and SGAN, with 0.9\% in the macro-F1 score. None of the tested semi-supervised methods registered a decrease in performance for this augmentation type.

The second best augmentation strategy was Manifold Mixup, which led to an average increase of 1.94\% in F1 score, with no semi-supervised method registering decreases. FixMatch's case had the biggest gains, with 2.23\% in macro F1, while MeanTeacher's case had the smallest increase, 0.62\%.

The third best augmentation method was paraphrasing. This method led to an average increase in F1 score by 1.6\%. This may be because the number of samples in the dataset increased, but it grew less than in the previous case of generating entirely new samples. Furthermore, the paraphrased samples are more similar to the original ones than the generated samples, thus bringing less diversification to the dataset. 

The fourth augmentation strategy was back-translation, which registered an average increase of 1.37\% in the F1 score. The only method that registered a slight decrease is LabelPropagation, which lost 0.04\% in macro-F1, while SGAN registered the most significant gain, with 4.30\%.

EDA was the only augmentation method that led to both improvements and decreases in performance. Also, the results remain largely the same with no augmentation applied. NoisyStudent presents the most significant increase with a 0.45\% gain in the macro-F1 score, while FixMatch registered the most notable decrease with a 0.96\% loss in the macro-F1 score. The reason for this is that the three augmentations (i.e., random insertion, random swap, and random deletion) used from EDA could change the meaning of a sentence. This could mean the sentence changes its offensive characteristic,  especially if a key offensive word was deleted.




\section{Conclusions}
This paper highlighted a selection of semi-supervised methods and data augmentation strategies used to improve the performance of classifiers on Romanian offensive comments. This particular context benefited from the added complexity of the evaluated techniques. This stems from the observed improvement in different classification metrics when the available data has more instances, even if unlabeled, compared to a default approach, which involves only fully supervised training and discarding the valuable information of the unlabeled samples. Thus, the RO-Offense dataset was presented and used in a novel context.
One possible direction of further research would be incorporating the evaluated semi-supervised methods into an ensemble context, where individual classifiers trained according to different techniques may boost the overall performance by specializing in various clusters of the input data.






\section*{Acknowledgements}
This work was supported by the NUST POLITEHNICA Bucharest through the PubArt program, and a grant from the National Program for Research of the National Association of Technical Universities - GNAC ARUT 2023.







\bibliography{k24-446}

\begin{thebibliography}{33}
\expandafter\ifx\csname natexlab\endcsname\relax\def\natexlab#1{#1}\fi
\providecommand{\url}[1]{\texttt{#1}}
\providecommand{\href}[2]{#2}
\providecommand{\path}[1]{#1}
\providecommand{\DOIprefix}{doi:}
\providecommand{\ArXivprefix}{arXiv:}
\providecommand{\URLprefix}{URL: }
\providecommand{\Pubmedprefix}{pmid:}
\providecommand{\doi}[1]{\href{http://dx.doi.org/#1}{\path{#1}}}
\providecommand{\Pubmed}[1]{\href{pmid:#1}{\path{#1}}}
\providecommand{\bibinfo}[2]{#2}
\ifx\xfnm\relax \def\xfnm[#1]{\unskip,\space#1}\fi
\bibitem[{Balestriero et~al.(2023)Balestriero, Ibrahim, Sobal, Morcos, Shekhar, Goldstein, Bordes, Bardes, Mialon, Tian et~al.}]{balestriero2023cookbook}
\bibinfo{author}{Balestriero, R.}, \bibinfo{author}{Ibrahim, M.}, \bibinfo{author}{Sobal, V.}, \bibinfo{author}{Morcos, A.}, \bibinfo{author}{Shekhar, S.}, \bibinfo{author}{Goldstein, T.}, \bibinfo{author}{Bordes, F.}, \bibinfo{author}{Bardes, A.}, \bibinfo{author}{Mialon, G.}, \bibinfo{author}{Tian, Y.}, et~al., \bibinfo{year}{2023}.
\newblock \bibinfo{title}{A cookbook of self-supervised learning}.
\newblock \bibinfo{journal}{arXiv preprint arXiv:2304.12210} .
\bibitem[{Berthelot et~al.(2019)Berthelot, Carlini, Goodfellow, Papernot, Oliver and Raffel}]{berthelot2019mixmatch}
\bibinfo{author}{Berthelot, D.}, \bibinfo{author}{Carlini, N.}, \bibinfo{author}{Goodfellow, I.}, \bibinfo{author}{Papernot, N.}, \bibinfo{author}{Oliver, A.}, \bibinfo{author}{Raffel, C.A.}, \bibinfo{year}{2019}.
\newblock \bibinfo{title}{Mixmatch: A holistic approach to semi-supervised learning}.
\newblock \bibinfo{journal}{Advances in neural information processing systems} \bibinfo{volume}{32}.
\bibitem[{Chen et~al.(2020)Chen, Yang and Yang}]{chen2020MixText}
\bibinfo{author}{Chen, J.}, \bibinfo{author}{Yang, Z.}, \bibinfo{author}{Yang, D.}, \bibinfo{year}{2020}.
\newblock \bibinfo{title}{Mixtext: Linguistically-informed interpolation of hidden space for semi-supervised text classification}, in: \bibinfo{booktitle}{Proceedings of the 58th Annual Meeting of the Association for Computational Linguistics}, pp. \bibinfo{pages}{2147--2157}.
\bibitem[{Chowdhury et~al.(2020)Chowdhury, Caragea and Caragea}]{chowdhury2020cross}
\bibinfo{author}{Chowdhury, J.R.}, \bibinfo{author}{Caragea, C.}, \bibinfo{author}{Caragea, D.}, \bibinfo{year}{2020}.
\newblock \bibinfo{title}{Cross-lingual disaster-related multi-label tweet classification with manifold mixup}, in: \bibinfo{booktitle}{Proceedings of the 58th annual meeting of the association for computational linguistics: student research workshop}, pp. \bibinfo{pages}{292--298}.
\bibitem[{Cojocaru et~al.(2022)Cojocaru, Paraschiv and Dascalu}]{cojocaru2022news}
\bibinfo{author}{Cojocaru, A.}, \bibinfo{author}{Paraschiv, A.}, \bibinfo{author}{Dascalu, M.}, \bibinfo{year}{2022}.
\newblock \bibinfo{title}{News-ro-offense-a romanian offensive language dataset and baseline models centered on news article comments}, in: \bibinfo{booktitle}{RoCHI}, pp. \bibinfo{pages}{65--72}.
\bibitem[{{Collins dictionary}()}]{collins}
\bibinfo{author}{{Collins dictionary}}, .
\newblock \bibinfo{title}{{Offensive language}}.
\newblock \bibinfo{howpublished}{https://www.collinsdictionary.com/dictionary/english/offensive-language}.
\newblock \bibinfo{note}{Online; accessed 07-Feb-2024}.
\bibitem[{Conneau et~al.(2020)Conneau, Khandelwal, Goyal, Chaudhary, Wenzek, Guzm{\'a}n, Grave, Ott, Zettlemoyer and Stoyanov}]{conneau2019unsupervised}
\bibinfo{author}{Conneau, A.}, \bibinfo{author}{Khandelwal, K.}, \bibinfo{author}{Goyal, N.}, \bibinfo{author}{Chaudhary, V.}, \bibinfo{author}{Wenzek, G.}, \bibinfo{author}{Guzm{\'a}n, F.}, \bibinfo{author}{Grave, E.}, \bibinfo{author}{Ott, M.}, \bibinfo{author}{Zettlemoyer, L.}, \bibinfo{author}{Stoyanov, V.}, \bibinfo{year}{2020}.
\newblock \bibinfo{title}{Unsupervised cross-lingual representation learning at scale}, in: \bibinfo{booktitle}{Proceedings of the 58th ACL}, \bibinfo{organization}{Association for Computational Linguistics}.
\bibitem[{Dumitrescu et~al.()Dumitrescu, Ilie and Kummervold}]{t5rohugg}
\bibinfo{author}{Dumitrescu, S.}, \bibinfo{author}{Ilie, M.}, \bibinfo{author}{Kummervold, P.E.}, .
\newblock \bibinfo{title}{{t5-v1\_1-base-romanian}}.
\newblock \bibinfo{howpublished}{{https://huggingface.co/dumitrescustefan/t5-v1\_1-base-romanian}}.
\newblock \bibinfo{note}{[Online; accessed 07-February-2024]}.
\bibitem[{Dumitrescu et~al.(2021)Dumitrescu, Rebeja, Lorincz, Gaman, Avram, Ilie, Pruteanu, Stan, Rosia, Iacobescu et~al.}]{dumitrescu2021liro}
\bibinfo{author}{Dumitrescu, S.D.}, \bibinfo{author}{Rebeja, P.}, \bibinfo{author}{Lorincz, B.}, \bibinfo{author}{Gaman, M.}, \bibinfo{author}{Avram, A.}, \bibinfo{author}{Ilie, M.}, \bibinfo{author}{Pruteanu, A.}, \bibinfo{author}{Stan, A.}, \bibinfo{author}{Rosia, L.}, \bibinfo{author}{Iacobescu, C.}, et~al., \bibinfo{year}{2021}.
\newblock \bibinfo{title}{Liro: Benchmark and leaderboard for romanian language tasks}, in: \bibinfo{booktitle}{Thirty-fifth Conference on Neural Information Processing Systems Datasets and Benchmarks Track (Round 1)}.
\bibitem[{Guo et~al.(2023)Guo, Lin, Liu, Zheng, Tu and Wang}]{guo2023implicit}
\bibinfo{author}{Guo, T.}, \bibinfo{author}{Lin, L.}, \bibinfo{author}{Liu, H.}, \bibinfo{author}{Zheng, C.}, \bibinfo{author}{Tu, Z.}, \bibinfo{author}{Wang, H.}, \bibinfo{year}{2023}.
\newblock \bibinfo{title}{Implicit offensive speech detection based on multi-feature fusion}, in: \bibinfo{booktitle}{International Conference on Knowledge Science, Engineering and Management}, \bibinfo{organization}{Springer}. pp. \bibinfo{pages}{27--38}.
\bibitem[{Hoefels et~al.(2022)Hoefels, {\c{C}}{\o}ltekin and M{\u{a}}droane}]{hoefels2022coroseof}
\bibinfo{author}{Hoefels, D.C.}, \bibinfo{author}{{\c{C}}{\o}ltekin, {\c{C}}.}, \bibinfo{author}{M{\u{a}}droane, I.D.}, \bibinfo{year}{2022}.
\newblock \bibinfo{title}{Coroseof-an annotated corpus of romanian sexist and offensive tweets}, in: \bibinfo{booktitle}{Proceedings of the Thirteenth Language Resources and Evaluation Conference}, pp. \bibinfo{pages}{2269--2281}.
\bibitem[{Hyun et~al.(2020)Hyun, Jeong and Kwak}]{hyun2020class}
\bibinfo{author}{Hyun, M.}, \bibinfo{author}{Jeong, J.}, \bibinfo{author}{Kwak, N.}, \bibinfo{year}{2020}.
\newblock \bibinfo{title}{Class-imbalanced semi-supervised learning}.
\newblock \bibinfo{journal}{arXiv preprint arXiv:2002.06815} .
\bibitem[{Iscen et~al.(2019)Iscen, Tolias, Avrithis and Chum}]{zhul2002label}
\bibinfo{author}{Iscen, A.}, \bibinfo{author}{Tolias, G.}, \bibinfo{author}{Avrithis, Y.}, \bibinfo{author}{Chum, O.}, \bibinfo{year}{2019}.
\newblock \bibinfo{title}{Label propagation for deep semi-supervised learning}, in: \bibinfo{booktitle}{Proceedings of the IEEE/CVF conference on computer vision and pattern recognition}, pp. \bibinfo{pages}{5070--5079}.
\bibitem[{Laine and Aila(2016)}]{laine2016temporal}
\bibinfo{author}{Laine, S.}, \bibinfo{author}{Aila, T.}, \bibinfo{year}{2016}.
\newblock \bibinfo{title}{Temporal ensembling for semi-supervised learning}.
\newblock \bibinfo{journal}{arXiv preprint arXiv:1610.02242} .
\bibitem[{Lee et~al.(2022)Lee, Kim, Kim, Cheon, Cho and Han}]{lee2022cr}
\bibinfo{author}{Lee, D.}, \bibinfo{author}{Kim, S.}, \bibinfo{author}{Kim, I.}, \bibinfo{author}{Cheon, Y.}, \bibinfo{author}{Cho, M.}, \bibinfo{author}{Han, W.S.}, \bibinfo{year}{2022}.
\newblock \bibinfo{title}{Contrastive regularization for semi-supervised learning}, in: \bibinfo{booktitle}{Proceedings of the IEEE/CVF Conference on Computer Vision and Pattern Recognition}, pp. \bibinfo{pages}{3911--3920}.
\bibitem[{Lee et~al.(2021)Lee, Shin and Kim}]{lee2021abc}
\bibinfo{author}{Lee, H.}, \bibinfo{author}{Shin, S.}, \bibinfo{author}{Kim, H.}, \bibinfo{year}{2021}.
\newblock \bibinfo{title}{Abc: Auxiliary balanced classifier for class-imbalanced semi-supervised learning}.
\newblock \bibinfo{journal}{Advances in Neural Information Processing Systems} \bibinfo{volume}{34}, \bibinfo{pages}{7082--7094}.
\bibitem[{Lu et~al.(2020)Lu, Du and Nie}]{lu2020vgcn}
\bibinfo{author}{Lu, Z.}, \bibinfo{author}{Du, P.}, \bibinfo{author}{Nie, J.Y.}, \bibinfo{year}{2020}.
\newblock \bibinfo{title}{Vgcn-bert: augmenting bert with graph embedding for text classification}, in: \bibinfo{booktitle}{Advances in Information Retrieval: 42nd European Conference on IR Research, ECIR 2020, Lisbon, Portugal, April 14--17, 2020, Proceedings, Part I 42}, \bibinfo{organization}{Springer}. pp. \bibinfo{pages}{369--382}.
\bibitem[{Masala et~al.(2020)Masala, Ruseti and Dascalu}]{masala2020robert}
\bibinfo{author}{Masala, M.}, \bibinfo{author}{Ruseti, S.}, \bibinfo{author}{Dascalu, M.}, \bibinfo{year}{2020}.
\newblock \bibinfo{title}{Robert--a romanian bert model}, in: \bibinfo{booktitle}{Proceedings of the 28th COLING 2020}, pp. \bibinfo{pages}{6626--6637}.
\bibitem[{Niculescu et~al.(2021)Niculescu, Ruseti and Dascalu}]{niculescu2021rogpt2}
\bibinfo{author}{Niculescu, M.A.}, \bibinfo{author}{Ruseti, S.}, \bibinfo{author}{Dascalu, M.}, \bibinfo{year}{2021}.
\newblock \bibinfo{title}{Rogpt2: Romanian gpt2 for text generation}, in: \bibinfo{booktitle}{2021 IEEE 33rd ICTAI}, \bibinfo{organization}{IEEE}. pp. \bibinfo{pages}{1154--1161}.
\bibitem[{Odena(2016)}]{odena2016semi}
\bibinfo{author}{Odena, A.}, \bibinfo{year}{2016}.
\newblock \bibinfo{title}{Semi-supervised learning with generative adversarial networks}.
\newblock \bibinfo{journal}{arXiv preprint arXiv:1606.01583} .
\bibitem[{Paraschiv et~al.(2023)Paraschiv, Ion and Dascalu}]{paraschiv2023offensive}
\bibinfo{author}{Paraschiv, A.}, \bibinfo{author}{Ion, T.A.}, \bibinfo{author}{Dascalu, M.}, \bibinfo{year}{2023}.
\newblock \bibinfo{title}{Offensive text span detection in romanian comments using large language models}.
\newblock \bibinfo{journal}{Information} \bibinfo{volume}{15}, \bibinfo{pages}{8}.
\bibitem[{Paraschiv et~al.()Paraschiv, Sandu, Cercel and Dascalu}]{paraschiv2022fighting}
\bibinfo{author}{Paraschiv, A.}, \bibinfo{author}{Sandu, I.}, \bibinfo{author}{Cercel, D.C.}, \bibinfo{author}{Dascalu, M.}, .
\newblock \bibinfo{title}{Fighting romanian offensive language with ro-offense: A dataset and classification models for online comments}.
\bibitem[{Park et~al.(2020)Park, Zhang, Jia, Han, Chiu, Li, Wu and Le}]{park2020noisy}
\bibinfo{author}{Park, D.S.}, \bibinfo{author}{Zhang, Y.}, \bibinfo{author}{Jia, Y.}, \bibinfo{author}{Han, W.}, \bibinfo{author}{Chiu, C.C.}, \bibinfo{author}{Li, B.}, \bibinfo{author}{Wu, Y.}, \bibinfo{author}{Le, Q.V.}, \bibinfo{year}{2020}.
\newblock \bibinfo{title}{Improved noisy student training for automatic speech recognition}.
\newblock \bibinfo{journal}{arXiv preprint arXiv:2005.09629} .
\bibitem[{Pires et~al.(2019)Pires, Schlinger and Garrette}]{pires2019multilingual}
\bibinfo{author}{Pires, T.}, \bibinfo{author}{Schlinger, E.}, \bibinfo{author}{Garrette, D.}, \bibinfo{year}{2019}.
\newblock \bibinfo{title}{How multilingual is multilingual bert?}, in: \bibinfo{booktitle}{Proceedings of the 57th ACL}, pp. \bibinfo{pages}{4996--5001}.
\bibitem[{Sohn et~al.(2020)Sohn, Berthelot, Carlini, Zhang, Zhang, Raffel, Cubuk, Kurakin and Li}]{sohn2020fixmatch}
\bibinfo{author}{Sohn, K.}, \bibinfo{author}{Berthelot, D.}, \bibinfo{author}{Carlini, N.}, \bibinfo{author}{Zhang, Z.}, \bibinfo{author}{Zhang, H.}, \bibinfo{author}{Raffel, C.A.}, \bibinfo{author}{Cubuk, E.D.}, \bibinfo{author}{Kurakin, A.}, \bibinfo{author}{Li, C.L.}, \bibinfo{year}{2020}.
\newblock \bibinfo{title}{Fixmatch: Simplifying semi-supervised learning with consistency and confidence}.
\newblock \bibinfo{journal}{Advances in neural information processing systems} \bibinfo{volume}{33}, \bibinfo{pages}{596--608}.
\bibitem[{Su et~al.(2023)Su, Li, Branco and Inkpen}]{su2023ssl}
\bibinfo{author}{Su, X.}, \bibinfo{author}{Li, Y.}, \bibinfo{author}{Branco, P.}, \bibinfo{author}{Inkpen, D.}, \bibinfo{year}{2023}.
\newblock \bibinfo{title}{Ssl-gan-roberta: A robust semi-supervised model for detecting anti-asian covid-19 hate speech on social media}.
\newblock \bibinfo{journal}{Natural Language Engineering} , \bibinfo{pages}{1--20}.
\bibitem[{Tarvainen and Valpola(2017)}]{tarvainen2017mean}
\bibinfo{author}{Tarvainen, A.}, \bibinfo{author}{Valpola, H.}, \bibinfo{year}{2017}.
\newblock \bibinfo{title}{Mean teachers are better role models: Weight-averaged consistency targets improve semi-supervised deep learning results}.
\newblock \bibinfo{journal}{Advances in neural information processing systems} \bibinfo{volume}{30}.
\bibitem[{Trandab{\u{a}}ț et~al.(2022)Trandab{\u{a}}ț, Gifu and Adrian}]{trandabat2022detecting}
\bibinfo{author}{Trandab{\u{a}}ț, D.}, \bibinfo{author}{Gifu, D.}, \bibinfo{author}{Adrian, P.}, \bibinfo{year}{2022}.
\newblock \bibinfo{title}{Detecting offensive language in romanian social media}.
\newblock \bibinfo{journal}{Procedia Computer Science} \bibinfo{volume}{207}, \bibinfo{pages}{2883--2890}.
\bibitem[{Verma et~al.(2019)Verma, Lamb, Beckham, Najafi, Mitliagkas, Lopez-Paz and Bengio}]{verma2019manifold}
\bibinfo{author}{Verma, V.}, \bibinfo{author}{Lamb, A.}, \bibinfo{author}{Beckham, C.}, \bibinfo{author}{Najafi, A.}, \bibinfo{author}{Mitliagkas, I.}, \bibinfo{author}{Lopez-Paz, D.}, \bibinfo{author}{Bengio, Y.}, \bibinfo{year}{2019}.
\newblock \bibinfo{title}{Manifold mixup: Better representations by interpolating hidden states}, in: \bibinfo{booktitle}{International conference on machine learning}, \bibinfo{organization}{PMLR}. pp. \bibinfo{pages}{6438--6447}.
\bibitem[{Wang et~al.(2022)Wang, Chen, Heng, Hou, Fan, Wu, Wang, Savvides, Shinozaki, Raj et~al.}]{Wang2022freematch}
\bibinfo{author}{Wang, Y.}, \bibinfo{author}{Chen, H.}, \bibinfo{author}{Heng, Q.}, \bibinfo{author}{Hou, W.}, \bibinfo{author}{Fan, Y.}, \bibinfo{author}{Wu, Z.}, \bibinfo{author}{Wang, J.}, \bibinfo{author}{Savvides, M.}, \bibinfo{author}{Shinozaki, T.}, \bibinfo{author}{Raj, B.}, et~al., \bibinfo{year}{2022}.
\newblock \bibinfo{title}{Freematch: Self-adaptive thresholding for semi-supervised learning}.
\newblock \bibinfo{journal}{arXiv preprint arXiv:2205.07246} .
\bibitem[{Wei and Zou(2019)}]{wei2019eda}
\bibinfo{author}{Wei, J.}, \bibinfo{author}{Zou, K.}, \bibinfo{year}{2019}.
\newblock \bibinfo{title}{Eda: Easy data augmentation techniques for boosting performance on text classification tasks}, in: \bibinfo{booktitle}{Proceedings of the 2019 Conference on EMNLP and the 9th IJCNLP}, pp. \bibinfo{pages}{6382--6388}.
\bibitem[{Wolf et~al.(2019)Wolf, Debut, Sanh, Chaumond, Delangue, Moi, Cistac, Rault, Louf, Funtowicz et~al.}]{wolf2019huggingface}
\bibinfo{author}{Wolf, T.}, \bibinfo{author}{Debut, L.}, \bibinfo{author}{Sanh, V.}, \bibinfo{author}{Chaumond, J.}, \bibinfo{author}{Delangue, C.}, \bibinfo{author}{Moi, A.}, \bibinfo{author}{Cistac, P.}, \bibinfo{author}{Rault, T.}, \bibinfo{author}{Louf, R.}, \bibinfo{author}{Funtowicz, M.}, et~al., \bibinfo{year}{2019}.
\newblock \bibinfo{title}{Huggingface's transformers: State-of-the-art natural language processing}.
\newblock \bibinfo{journal}{arXiv preprint arXiv:1910.03771} .
\bibitem[{Zhang et~al.(2017)Zhang, Cisse, Dauphin and Lopez-Paz}]{zhang2017mixup}
\bibinfo{author}{Zhang, H.}, \bibinfo{author}{Cisse, M.}, \bibinfo{author}{Dauphin, Y.N.}, \bibinfo{author}{Lopez-Paz, D.}, \bibinfo{year}{2017}.
\newblock \bibinfo{title}{mixup: Beyond empirical risk minimization}.
\newblock \bibinfo{journal}{arXiv preprint arXiv:1710.09412} .

\end{thebibliography}
\bibliographystyle{elsarticle-harv}



\clearpage
\normalMode




\end{document}